\documentclass[conference]{IEEEtran}
\IEEEoverridecommandlockouts
\usepackage{cite}

\usepackage{amsmath,amssymb,amsfonts}
\usepackage{algorithmic}
\usepackage{graphicx}
\usepackage{xcolor}
\usepackage{booktabs}
\usepackage{float}
\usepackage[colorlinks=false,pdfborder={0 0 0}]{hyperref}
\usepackage{cleveref}
\usepackage{multirow}

\usepackage{url}

\def\BibTeX{{\rm B\kern-.05em{\sc i\kern-.025em b}\kern-.08em
    T\kern-.1667em\lower.7ex\hbox{E}\kern-.125emX}}

\begin{document}

\title{Learning Actionable World Models\\ for Industrial Process Control\\
\thanks{This work is supported by Innosuisse grant 62174.1 IP-ENG ``DISTRAL''.}
}

\author{
    \IEEEauthorblockN{\href{https://orcid.org/0009-0006-0236-4707}{\includegraphics[scale=0.08]{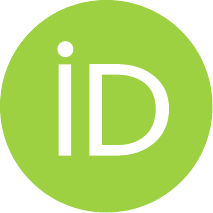}}\hspace{1mm}Peng Yan\IEEEauthorrefmark{1}\IEEEauthorrefmark{2}, 
    \href{https://orcid.org/0000-0003-4679-8081}{\includegraphics[scale=0.08]{pic/orcid.pdf}}\hspace{1mm}Ahmed Abdulkadir\IEEEauthorrefmark{1},
    \href{https://orcid.org/0009-0002-5760-9346}{\includegraphics[scale=0.08]{pic/orcid.pdf}}\hspace{1mm}Gerrit A. Schatte\IEEEauthorrefmark{3},
    \href{https://orcid.org/0000-0001-6759-5270}{\includegraphics[scale=0.08]{pic/orcid.pdf}}\hspace{1mm}Giulia Aguzzi\IEEEauthorrefmark{3},
    \href{https://orcid.org/0000-0003-1942-3866}{\includegraphics[scale=0.08]{pic/orcid.pdf}}\hspace{1mm}Joonsu Gha\IEEEauthorrefmark{2},
    \href{https://orcid.org/0009-0005-9348-2137}{\includegraphics[scale=0.08]{pic/orcid.pdf}}\hspace{1mm}Nikola Pascher\IEEEauthorrefmark{1},
    \\
    \href{https://orcid.org/0000-0002-7577-783X}{\includegraphics[scale=0.08]{pic/orcid.pdf}}\hspace{1mm}Matthias Rosenthal\IEEEauthorrefmark{1},
    \href{https://orcid.org/0000-0003-2843-9878}{\includegraphics[scale=0.08]{pic/orcid.pdf}}\hspace{1mm}Yunlong Gao\IEEEauthorrefmark{5},
    \href{https://orcid.org/0000-0001-8560-2120}{\includegraphics[scale=0.08]{pic/orcid.pdf}}\hspace{1mm}Benjamin F. Grewe\IEEEauthorrefmark{2}, and
    \href{https://orcid.org/0000-0002-3784-0420}{\includegraphics[scale=0.08]{pic/orcid.pdf}}\hspace{1mm}Thilo Stadelmann\IEEEauthorrefmark{1}\IEEEauthorrefmark{2}\IEEEauthorrefmark{6}}
    \IEEEauthorblockA{\{yanp,abdk,pash,rosn,stdm\}@zhaw.ch, \{giulia.aguzzi,gerrit.schatte\}@kistler.com, gaoyl@xmu.edu.cn, \{joogha,bgrewe\}@ethz.ch}
    \IEEEauthorblockA{\IEEEauthorrefmark{1}School of Engineering, Zurich University of Applied Sciences, Winterthur, Switzerland}
    \IEEEauthorblockA{\IEEEauthorrefmark{2}Institute of Neuroinformatics, ETH Zurich \& University of Zurich, Zurich, Switzerland}
    \IEEEauthorblockA{\IEEEauthorrefmark{3}Kistler Instrumente AG, Winterthur, Switzerland}
    \IEEEauthorblockA{\IEEEauthorrefmark{5}Pen-Tung Sah Institute of Micro-Nano Science and Technology, Xiamen University, Xiamen, China} 
    \IEEEauthorblockA{\IEEEauthorrefmark{6}Fellow, ECLT (European Centre for Living Technology, Venice, Italy) and Senior Member, IEEE}
}

\maketitle

\begin{abstract}
To go from (passive) process monitoring to active process control, an effective AI system must learn about the behavior of the complex system from very limited training data, forming an ad-hoc ``digital twin'' with respect to process inputs and outputs that captures the consequences of actions on the process's ``world''.
We propose a novel methodology based on learning world models that disentangles process parameters in the learned latent representation, allowing for fine-grained control.
Representation learning is driven by the latent factors influencing the processes through contrastive learning within a joint embedding predictive architecture.
This makes changes in representations predictable from changes in inputs and vice versa, facilitating interpretability of key factors responsible for process variations, paving the way for effective control actions to keep the process within operational bounds.
The effectiveness of our method is validated on the example of plastic injection molding, demonstrating practical relevance in proposing specific control actions for a notoriously unstable process.



\end{abstract}





\section{Introduction}


\begin{figure}[!th]
\centering
\includegraphics[width=\columnwidth]{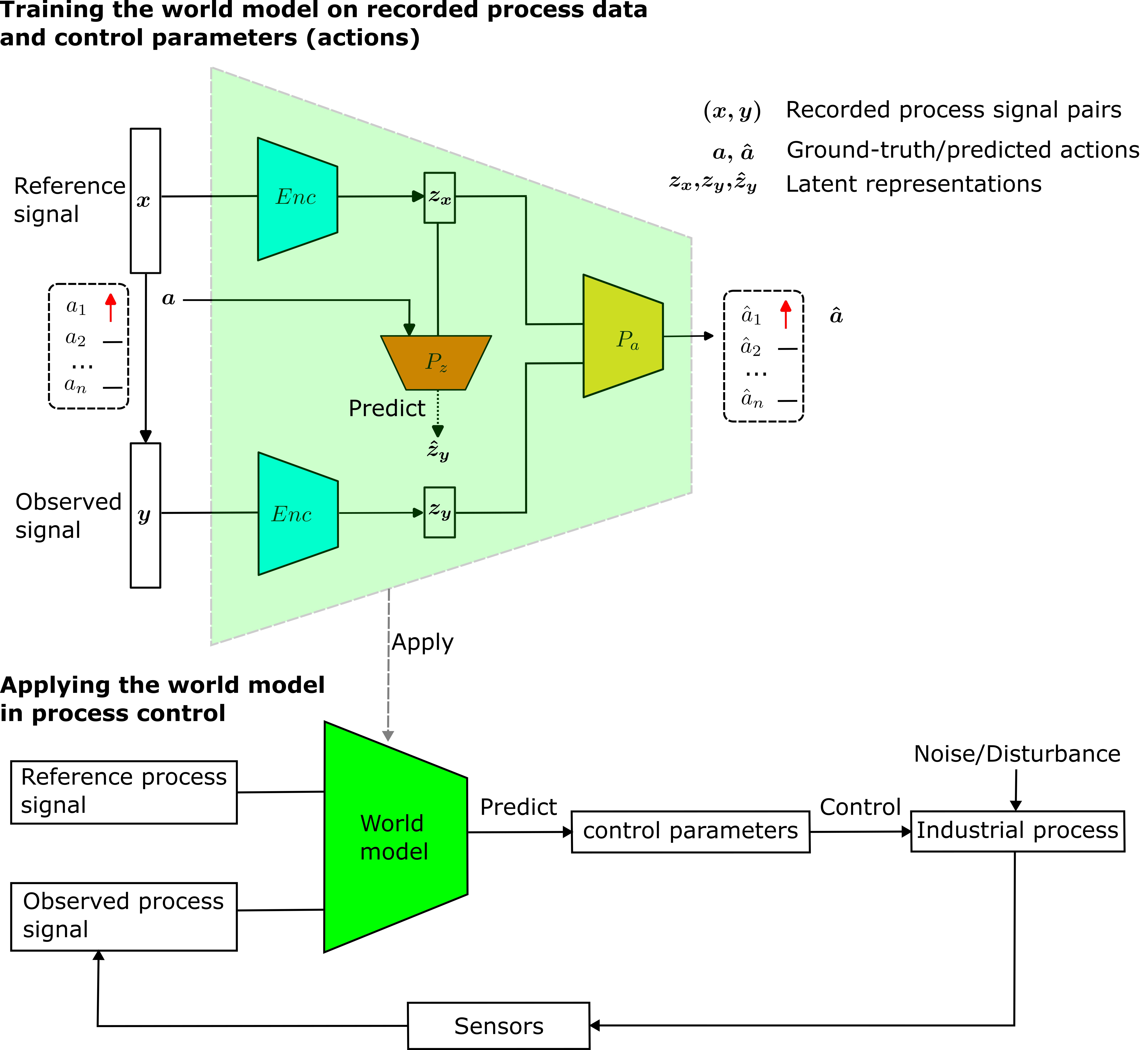}
\caption{Overview of world model training and application for process control. \emph{Top}: World model training using recorded process signal pairs (\(\boldsymbol{x}, \boldsymbol{y}\)) that have been generated in the same environment (world) but with changed control parameters (actions) \(\boldsymbol{a}=a_1,...,a_n\). The encoder (\(Enc\)) generates latent representations (\(\boldsymbol{z_x},\boldsymbol{z_y}\)) used for predicting the actions \(\hat{\boldsymbol{a}}\)
through the action prediction module (\(P_a\)). 
The latent predictive module (\(P_z\)) guides the learning of the latent representation by incorporating actions, ensuring the representations \(\boldsymbol{z}_x\) and \(\boldsymbol{z}_y\) align with the process's operational dynamics for more accurate and interpretable predictions.
Detailed modeling and training is discussed in Sec.~\ref{method}. \emph{Bottom}: Application of the trained world model in real-time process control where the model predicts actions based on a reference process signal plus an observed process signal to maintain stability despite changing external conditions or influences.
}
\label{fig:overall}
\end{figure}

Automatic process monitoring plays a central role in maintaining repeatable and stable processes with high-quality products in industrial processes \cite{schoningAIClosedLoopControl2022}.
Commonly, the monitoring task is formulated as an anomaly detection task, as it is \textsl{a priori} unknown how exactly a process may deviate from normal operations \cite{yan_comprehensive_2024}.
To restore the system to its normal operational status once an anomaly is detected, it is necessary to identify and remove the cause of any detected anomaly.
Existing approaches focus on anomaly detection \cite{yanSDS2024} and lack interpretability, making it difficult to identify the underlying root causes.
One reason is that real-world monitored processes---such as plastic injection molding or air condition flow---often lack an analytical or simulation model of the relationship between measured process signals and control parameters (actions: physical signals, such as mold temperature adjustments, directly applied to the system, e.g., an injection molding machine, to regulate its behavior).
This complicates root cause analysis for control despite reliable algorithms for anomaly detection.
Thus, understanding relations between process signals and actions is crucial to transition from passive monitoring to active control. 

Machine learning (ML) systems have shown large potential to predict process signal drifts or disturbances before they impact operations, potentially replacing conventional controller blocks by directly interpreting sensor values as process feedback \cite{Schatte2023}. 
To become fully effective for process \emph{control} in complex, dynamic processes, \emph{world models} appear as a promising solution. Having become a branch of ML research with notable progress in recent years \cite{ha_world2018, lecun2022path, wu2023daydreamer}, they are typically referred to as a learned representation of the physical world, capable of predicting the consequences of various actions on the world’s state \cite{garrido2024learningleveragingworldmodels}. While many world models are built as generative models of specific environments (``world'' referring to any relevant part of the environment) to serve as simulations for AI agent training (such as \cite{ha_world2018,wu2023daydreamer}), we are concerned here with \emph{actionable} world models that learn consequence-aware representations of world states (process signals, e.g., time-series data) 
 and directly output control actions (e.g., scalars). 

The integration of such world models into process monitoring and control systems presents a significant opportunity to advance intelligent, adaptive, and highly efficient industrial operations. Fig.~\ref{fig:overall} illustrates our proposed approach for process control, where the world model is trained on recorded process signal pairs and their control parameters and can later be applied to a running process to bring it back to normal operation if it deviates from a reference signal. This approach enables real-time adjustments to control parameters based on incoming process signals, ensuring stability and consistency in the presence of anomalies such as noise or disturbances. 
That the world model's training integrates actions and states enriches the system's grasp of complex dynamics, enabling robust performance across varying operational conditions. 

In this paper, we introduce a novel world model for process control, inspired by joint embedding predictive architectures (JEPAs) and contrastive learning \cite{lecun2022path, le2020contrastive}.
Our world model makes use of recorded process signals and corresponding actions to learn an action-aware latent representation, facilitating disentangled action prediction in the presence of many possible control parameter adjustments. 
Experimental results on a real-world use case from plastic injection molding control demonstrate that our world model achieves its goals with high sample efficiency (limited industrial data of only 80 samples).
%
%
Our main contributions are:
(a) a world model for process control, based on a novel architecture, inspired by the JEPA framework and contrastive learning; it conditions the learned latent representation of the world's state on the difference in parameters evoking this state as compared to a reference state;
(b) experimental validation on a real-world injection molding case, where the contrastive modeling of pressure curve pairs and corresponding machine parameters allows the model to better distinguish between subtle variations in the data, showing disentanglement for different actions in the embedding space for the first time as compared to the state of the art;
(c) a novel evaluation scheme for the quality of disentanglement, based on angle and distance deviations between ground truth and predicted actions and showing the validity of our approach for real-world process control.

\section{Background and related work}




\emph{Contrastive learning.}
In this form of self-supervised learning (SSL) \cite{hadsellDimensionalityReductionLearning2006}, a model is trained to bring the embeddings of a sample and a transformed version thereof close to each other, while non-corresponding samples are pushed away from each other. Early work includes Siamese networks \cite{bromleySignatureVerificationUsing1993}. Contrastive methods have been introduced for time-series analysis, e.g., \cite{eldeleTimeSeriesRepresentationLearning2021d, ozyurtContrastiveLearningUnsupervised2023}, including work that disentangles variant and invariant factors in a pair of observations \cite{hamaguchiRareEventDetection2019}. 

Typically, contrastive learning relies heavily on large numbers of contrastive pairs, particularly negatives, to learn the decision boundaries \cite{lecun2022path}. We take the intuition of contrastive learning by learning from process signal pairs, but inform them with the action information from the differences in the corresponding control parameters. This allows our model to better recognize and differentiate the nuanced effects of control parameters with respect to variations in the process signal.
We further reduce the dependency on extensive contrastive pairs by leveraging action-aware embeddings. 

\emph{Generative and world modeling.}
Variational Auto-Encoders (VAEs) learn regularized low-dimensional latent representation by using a encoder-decoder architecture.
Yan et al. adapt a convolutional VAE to learn a general representation of industrial time-series data, and add additional layers to predict corresponding control (machine) parameters \cite{yanSDS2024}. The model is good in detecting anomalies by identifying data drift. However, the learned representations cannot reliably disentangle the factors that affect the time-series. Our work aims to improve on this.
Keurti et al. propose the homomorphism auto-encoder, equipped with a group representation acting on its latent space to learn internal representations of sensory information that are consistent with actions that modify it \cite{keurti2023homomorphism}. 

This actionable representation represents a form of world modeling. In general, world modeling approaches can be distinguished by the kind of representations they learn. Monolithic latent representations feature a single latent vector jointly representing both states and actions \cite{ha_world2018, hafner2023mastering, lin2023learning, bruce2024genie, alonso2024diffusion}. Structured representations however can offer higher sample efficiency, out-of-distribution efficiency and interpretability. Among such approaches are those that allow compositionality with respect to spatial \cite{biza2022binding} or temporal \cite{gumbsch2023learning} factors. Learning structured representations for action spaces is challenging due to the combinatorial nature of actions (one can, e.g., go `up' and `left' at the same time). Hence, world models often fall back on monolithic action space representations \cite{chandak2019learning, tennenholtz2019natural}. We take this approach as well.

\emph{Joint embedding representation learning.}
The quality of input reconstruction from a generative model does not necessarily correspond to the quality of the learned representations \cite{chen2024deconstructingdenoisingdiffusionmodels}. JEPAs have emerged as a promising SSL approach for representation learning, offering an alternative to generative models: Instead of taking a loss in the reconstructed space, they predict and compute the loss in latent space, conditioned on a latent variable (e.g., actions) \cite{lecun2022path}. 
Recent JEPA variants (I-JEPA \cite{assran2023selfsupervisedlearningimagesjointembedding} and V-JEPA \cite{bardes2024revisitingfeaturepredictionlearning}) predict masked parts of the input, learning representations for computer vision tasks.
We extend the JEPA framework by moving away from an implicit masking action ($\boldsymbol{x}\xrightarrow{\boldsymbol{masking}}\boldsymbol{y}$) to explicitly providing arbitrary actions in a factored form ($\boldsymbol{x}\xrightarrow{\boldsymbol{a}}\boldsymbol{y}$), facilitating learning a rich latent representation of cause and effect economically.
%
%
%
%
%
%
%
Thus, we extend the typical SSL framework by not being agnostic to external factors (like machine parameters). 

Instead, our approach is \emph{action-aware} self-supervised learning, combining the strengths of self-supervised feature extraction with task-specific guidance from actions (control parameters).
Our world model uses them to directly guide representation learning, ensuring that the learned embeddings are actionable. 
This avoids the need for contrastive pairs or negative sampling strategies, reducing computational overhead and simplifying the training process.
Control parameters (e.g. machine parameters in injection molding) provide explicit domain knowledge, making the representations interpretable and well-suited for process control. 

%
Furthermore, the supervisory signal through action conditioning in our approach naturally prevents the collapse of the learned representations, eliminating the need for additional regularization techniques as otherwise common in SSL. 

\emph{Disentangled representation learning for time-series.}
Deep learning has been proposed to learn effective representations of the time-series data for downstream applications \cite{maLearningRepresentationsTime2019, stadelmann2019beyond, eldeleTimeSeriesRepresentationLearning2021d, trirat_universal_2024}. To be useful for fine-grained process control, the learned representations must be disentangled: A low-dimensional latent representation \( \boldsymbol{z} = \{z_1, z_2, \ldots, z_n\} \) is sought such that all pairs of latent variables in \( Z \) are independent, i.e., the change of the sequential patterns in the input corresponding to variable $z_i$ is invariant to variable $z_j$, denoted as  \( z_i \perp \!\!\! \perp z_j \), \(\forall i, j\), where \( i,j \in \{1, 2, \ldots, n\} \) and \( i \ne j \). The disentanglement can be extended to group disentanglement so that a group of latent variables is independent of other different groups. Most representation learning relies on unsupervised learning due to label scarcity. However, unsupervised learning of disentangled representations is fundamentally impossible without inductive biases on both the models and the data \cite{locatelloChallengingCommonAssumptions2019}. 
While interpretable semantic concepts often depend on the interaction of multiple factors rather than individual components, Li et al. argue that it is essential to interpret time-series data through a single latent representation in industrial applications \cite{liLearningDisentangledRepresentations2021}. 
A disentangled representation can be decomposed into separate dimensions, 
where the change in each dimension reflects a distinctive change in the real world. 
Li et al. propose to learn interpretable time-series representations through encouraging high mutual information between the latent codes and the original time-series to achieve disentangled latent representations \cite{liLearningDisentangledRepresentations2021}. Hamaguchi et al. \cite{hamaguchiRareEventDetection2019} use contrastive learning to learn common and specific features during representation learning. Then, only the common features are used for downstream tasks. 
In contrast, we address disentanglement through action-conditioned world modeling.

\section{Methodology} \label{method}



\begin{figure}[t]
\centering
\includegraphics[width=0.9\columnwidth]{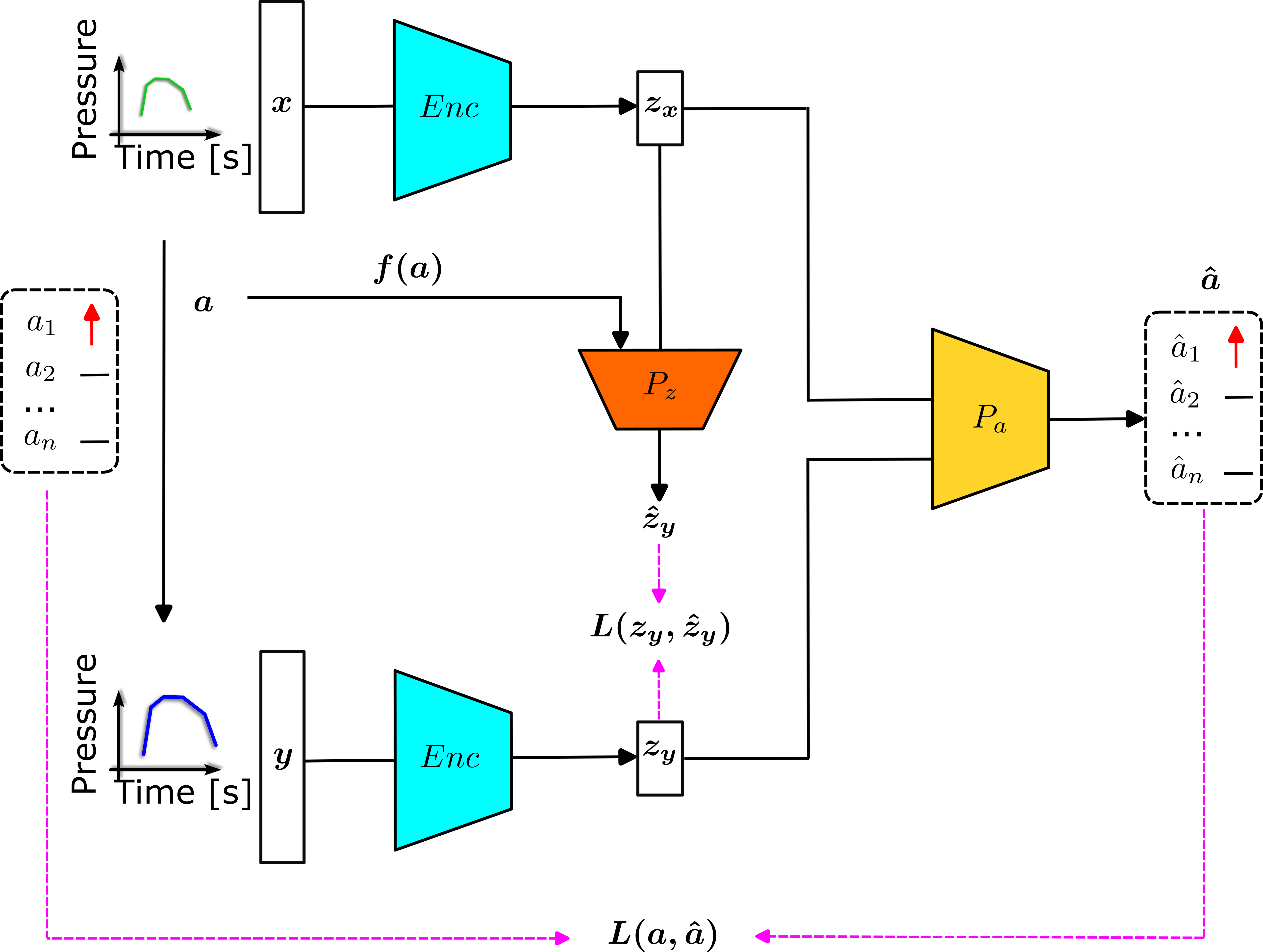}
\caption{Proposed actionable world model architecture (general case; inputs for injection molding example shown, see Sec.~\ref{sec:application}).
There are $3$ main components in the model, encoder $Enc$, latent predictor $P_z$, and action predictor $P_a$. Losses are computed in representation space and for the action prediction.}
\label{fig_model}
\end{figure}

Given a dataset consisting of process signals (reference $\boldsymbol{x}$, observation $\boldsymbol{y}$), and the corresponding actions \(\boldsymbol{a} = [ a_1, a_2, ..., a_n]\), the objective of the modeling is to predict the action that is responsible for
transforming the reference signal into the observed one.
To achieve this, the machine must build a world model.
We implement the world model using the deep learning architecture depicted in
Fig.~\ref{fig_model}.
The elements of the architecture are detailed below.

\emph{Encoder.}
The two encoders with shared weights map the reference signal \(\boldsymbol{x} = (x_1, \dots, x_n)\) and the observed signal \(\boldsymbol{y} = (y_1, \dots, y_n)\) into latent representations \(\boldsymbol{z_x}\) and \(\boldsymbol{z_y}\), respectively.
\(\boldsymbol{z_x}\) and \(\boldsymbol{z_y}\) capture the essential features of the process signal and the process's dynamics.

\emph{Latent predictor.}
The latent predictor \(P_z\) takes as input the concatenation of the latent representation \(\boldsymbol{z_x}\) and 
the transformed action $f(\boldsymbol{a})$. In our implementation, \(f(\cdot)\) is simply the identity function. 

\(P_z\) outputs $\boldsymbol{\hat{z}_y}$, the prediction of the representation $\boldsymbol{z_y}$ that follows from repeating the process (that led to $\boldsymbol{x}$) after applying the change in control parameters ($\boldsymbol{a}$), modeling the transition between states caused by an action: $\boldsymbol{x}\xrightarrow{\boldsymbol{a}}\boldsymbol{y}$.

\emph{Action predictor.}
The action predictor $P_a$ takes the difference between the latent representations of the process signals ($\boldsymbol{z_y} - \boldsymbol{z_x}$) as input and predicts the corresponding control action $\boldsymbol{\hat{a}}$ that induced the observed change. Formally, this is expressed as: \(\boldsymbol{{\hat a}} = P_a(\boldsymbol{z_y} - \boldsymbol{z_x})\).



To train the model effectively and improve the disentanglement of action dimensions in the latent space, we define a loss that integrates multiple components, as depicted in Fig.~\ref{fig_model}.

\emph{Latent consistency loss. }
It enforces consistency between the predicted latent representation \(\boldsymbol{\hat{z}_y}\) and the encoded representation \(\boldsymbol{z_y}\). This is achieved by minimizing the distance between these two representations, which can be expressed as:
\begin{equation}
    L(\boldsymbol{z_y}, \boldsymbol{\hat{z}_y}) = ||\boldsymbol{z_y} - \boldsymbol{\hat{z}_y}||^2
\end{equation}
\emph{Action prediction loss.}
It penalizes the discrepancy between the predicted action
\(\hat{\boldsymbol{a}}\),
and the action \(\boldsymbol{a}\) that is actually applied. This is critical for disentangling the effect of each action dimension. The loss is defined as:
\begin{equation}
    L(\boldsymbol{a}, \hat{\boldsymbol{a}}) = \|
    \boldsymbol{a} - \hat{\boldsymbol{a}}\|^2
\end{equation}
\emph{Overall objective.}
The complete loss function is a linear combination of the above components:
\begin{equation}
    \mathcal{L}_{\text{total}} = \lambda_1 L(\boldsymbol{z}_y, \boldsymbol{\hat{z}}_y) + \lambda_2 L(\boldsymbol{a}, \hat{\boldsymbol{a}})
\end{equation}
where \(\lambda_1\) and \(\lambda_2\) are hyper-parameters that balance the contribution of each loss term.

By jointly optimizing both objectives, the model learns action-aware latent representations guided by observing combinations of action and corresponding signals.
Eventually, it learns to predict the effect of a continuum of actions on the signal.
This framework ensures consistency in latent space, disentangled action dimensions, and supports downstream tasks like anomaly detection and domain shift identification besides process control.
\section{Application to injection molding}
\label{sec:application}

\subsection{Use case}
\emph{Plastic injection molding.} This is a widely utilized industrial process in which granular plastic is melted, injected into a mold, and cooled to produce components for various applications \cite{zhaoIntelligentInjectionMolding2020}. The repetitive nature of this process makes it particularly suitable for automation, but process monitoring and control is essential to ensure high product quality. 
Pressure sensors installed in mold cavities collect the pressure footprint of manufactured parts over production time, providing critical insight into process and product quality while enabling detection of potential anomalies \cite{Vaculik2020}. Although systems exist that automatically detect such anomalies \cite{kistler2023comoneo}, identifying their root causes remains a significant challenge \cite{yanSDS2024}. Root causes are difficult to replicate in a controlled experimental setup. However, their effect on pressure signals can be mimicked by adjusting machine parameters, such as injection speed or holding pressure, which influence curve shape, scale and time shifting. 
For example, excess of plastic in a produced part that distorts its shape and reduces its quality, also known as flash defect, may appear as a pressure curve with higher peak than the original stable process, similar to the effect of increasing holding pressure \cite{Vaculik2020}. Therefore, predicting variations in these parameters can help technicians quickly identify potential sources of anomalies and resolve them. Full automatic process control in a closed-loop fashion is however not possible in this setting for the reason that (staying in the aforementioned example) if the model predicts that pressure needs to be reduced, the correct control action is not to reduce the pressure (adjust the machine parameter), but to clean the sensor.

\emph{World modeling.} We implement our methodology by adopting the deep learning architecture illustrated in Fig.~\ref{fig_model} and detailed in
Sec.~\ref{method} to an injection molding use case.
The observed signals are transient pressure curves (time-series). The action space represents changes in machine parameters with respect to a reference setting (vector with a scalar per parameter).
It is important to note that in this framework, the \textsl{world} is specific to each product being produced, as the mold, the location of the sensor, and other variables impact the signal.
Thus, the model must learn an ad-hoc representation of its specific \textsl{world} for each different
production setting, based on a few observations. It may rely on transfer learning from previous productions to deal with data needs economically.

\subsection{Datasets}
\label{sec:data}

\begin{table}[t]
\caption{Pre-training/Fine-tuning dataset characteristics}
\begin{center}
\begin{tabular}{|c|c|c|c|c|}
\hline
\specialrule{0.6pt}{0pt}{0pt} 
\textbf{Dataset} & \textbf{N}&  \textbf{Cycle duration [s]}  & \textbf{No. of actions}  & \textbf{Action dim.} \\

\hline \hline
D1  & $270$   &	$10$ & $27$ & $3$\\
\hline
D2  & $270$   &	$10$ & $27$ & $3$\\
\hline
D3   & $460$   &	$20$ & $23$ & $6$\\
\hline
\specialrule{0.6pt}{0pt}{0pt} 

\end{tabular}
\label{cal_test}
\end{center}
\end{table}

\emph{Data collection.}
We collect pressure signals from injection molding processes, following a methodology similar to \cite{yanSDS2024},  using Kistler cavity pressure sensors. Details of the $3$ datasets can be seen in Tab.~\ref{cal_test}.

Datasets D1 and D2 differ in the specific products (plastic parts) produced. Otherwise, we follow the same systematic design of experiments (DOE), in which we 
sparsely sample signals obtained by varying a small number of machine parameters, namely holding pressure, injection speed, and mold temperature.
%
%
Specifically, we systematically set each machine parameter to one of $3$ levels, resulting in a total of $3^3 = 27$ unique parameter combinations. 
We record $10$ production cycles for each machine parameter combination after reaching a steady production state, yielding in total $27 \times 10=270$ samples in each of these two datasets. The $10$ cycles per parameter combination are expected to yield almost similar pressure curves during stable production conditions, hence serving as a form of data augmentation of the $27$ fundamentally different settings.
An input to the model $(\boldsymbol{x}, \boldsymbol{y}, \boldsymbol{a})$ consists of
a reference signal \(\boldsymbol{x}\), an observation \(\boldsymbol{y}\), and the action \(\boldsymbol{a}\) (the difference between the machine parameters of the observed and the reference signal). 

\begin{figure}[!ht]
    \centering
    \includegraphics[width=0.4\textwidth]{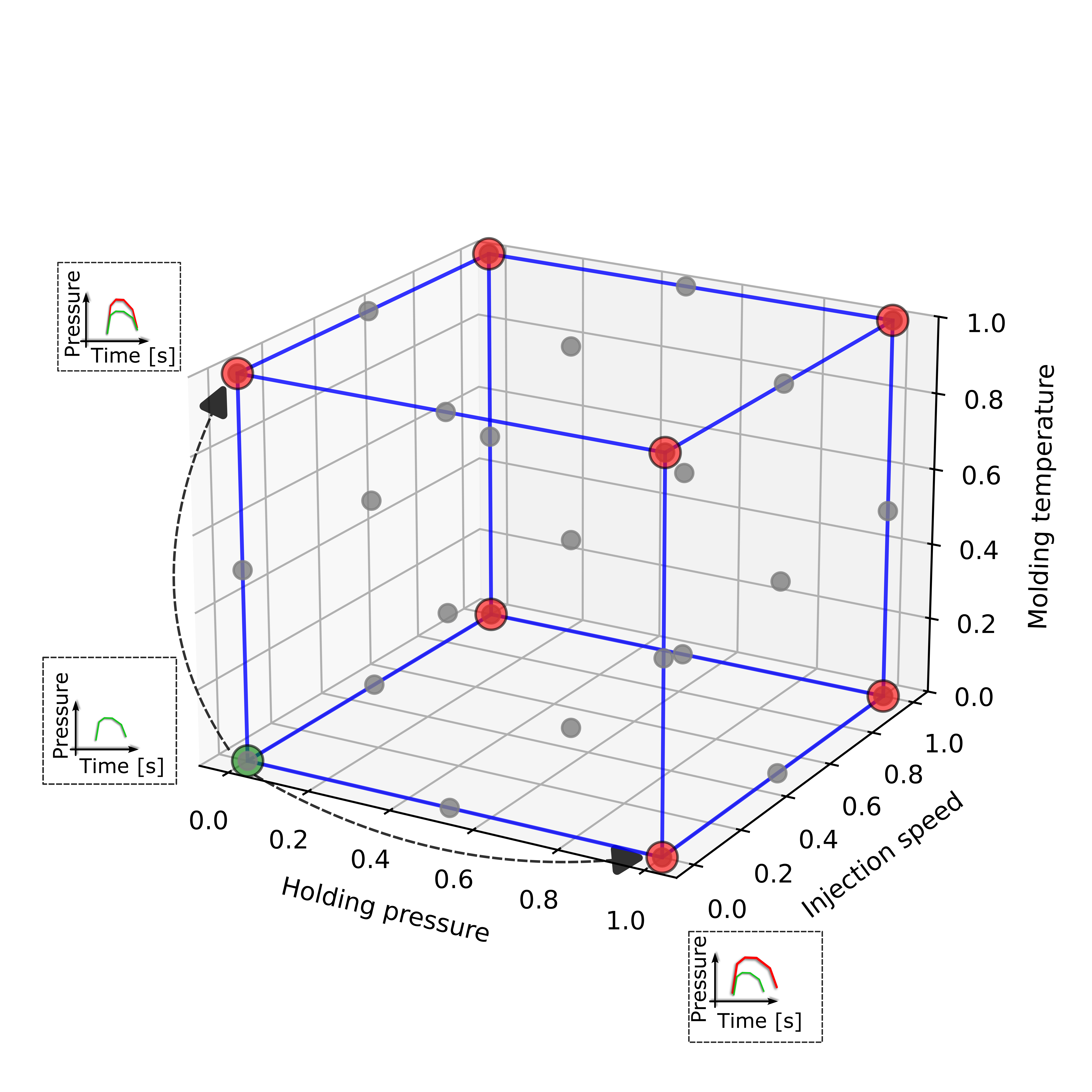}
    \caption{Illustration of the three-dimensional machine parameter space of D1/D2 with a normalized scale. 
    The gray points represent $27$ different machine parameters settings. 
    The green and red points represent the most extreme settings in the corners of the cube (outline indicated by blue lines for clarity), in which individual parameters take on either minimal or maximal values.
    The pressure curves shown in the dashed rectangles show one example of a 
    reference signal $\boldsymbol{x}$ (green) and two observed signal $\boldsymbol{y}$ (red), each corresponding to different machine parameter settings; the arrows represent an action to transition from $\boldsymbol{x}\xrightarrow{\boldsymbol{a}}\boldsymbol{y}$.
    }
    \label{fig:cube}
\end{figure}

Fig.~\ref{fig:cube} illustrates the DOE behind D1 and D2. Visualizing the 3D structure of the resulting machine parameter space helps in comprehending the considerations behind dataset construction. 


Dataset D3 represents data from the production of a third product and includes $3$ more machine parameters in addition to the previous $3$: hot runner temperature, dosing speed, and holding time. The parameters vary less systematically, resembling more a collection ``in the wild'' as it would happen during normal operations.

This results in a total of $46$ different actions, with $20$ cycles repeated per setting. By carefully analyzing the resulting pressure curves manually, we filter out $23$ of these actions that do not result in a stable process (i.e., their $20$ cycles vary significantly). Overall, this yields $23 \times 20=460$ pressure curves (that are primarily explained by varying the original $3$ parameters, with the additional $3$ playing a noise-like function for our purposes).

\emph{Data preprocessing.}
The pressure time-series represent the dynamic changes in pressure over time, which are critical for understanding the process's behavior. As the original time-series data is sampled with irregular time intervals, we preprocess the data by resampling the data uniformly to $500$ time points between $0$ and $10$ seconds. This interpolation ensures uniformity across all pressure curve samples.
Additionally, we normalize pressure data by dividing by $1000$,
roughly rescaling the observed data to a range between $0$ and $1$. 

\subsection{Training details} 
The encoder \(Enc\) consists of $6$ convolutional layers and $1$ fully connected layer (FCL), outputting a $10d$ latent space. The latent predictor \(P_z\) consists of a single FCL, mapping $13d$ input to $10d$ output, while the action predictor \(P_a\) maps the $10d$ latent space into $3d$ using a single FCL with zero bias, so its weights can directly represent the relation between latent variables and actions, facilitating disentanglement analysis.

We train the world model using stochastic gradient descent with batch size $32$ for $500$ epochs using the Adam optimizer \cite{KingBa15} with learning rate $\alpha=3 \times 10^{-4}$ and the default settings otherwise.
For fine-tuning, the model is first pre-trained for $10$ epochs. Then, all parameters of \(P_z\), \(P_a\) and the last layer of $Enc$ are fine-tuned for $500$ epochs. 
The coefficients for the loss function in Eq.~(4) are set as 
$\lambda_1 = 1, \lambda_2 = 10$, balancing the terms such that they are of roughly equal magnitude.

\section{Experimental setup and evaluation}
\label{sec_experiments}

\subsection{Experiments}
We design $4$ experiments to assess the ability of our world model to suggest process control actions in a real-world setting, trained in an economically viable fashion. 

\emph{Experiment 1: Action space coverage.} The goal is to learn to what extent the world model can learn the effects of actions across \emph{the whole} action space, based on a sparse sampling of all possible actions. We conduct this analysis individually on datasets D1 and D2 that offer a systematic coverage of the action space.
We use as training data the samples corresponding with the machine parameters in the $8$ corners of the cube of Fig.~\ref{fig:cube}. 
Specifically, we pair each of the $8 \times 10$ samples with every other, resulting in $6400$ training samples.
For testing, we take the remaining $19 \times 10$ samples (representing the inner area of the cube and thus allowing a comprehensive analysis) and pair them in turn with each sample corresponding with machine parameters from the $8$ corners as  references. This results in $(19 \times 10) \times (8 \times 10)=15200$ test inputs.
The use of training data in testing here does not constitute data leakage, as the references (that are used in training data) are an inherent ingredient of the world model
as the observed signal that is not included in the training is used to predict the actions \emph{in reference to $\boldsymbol{x}$}.
We average over the results per individual reference vertex to calculate our evaluation metrics. Additionally, to ensure reliable results, we repeat each experiment $6$ times with different random seeds for model initialization and finally report the mean value derived from these evaluations.

\emph{Experiment 2: Sample efficiency.} We are interested in learning the model's performance when only granted minimal training data, considering the time and cost constraints for the DOE in industry. Specifically, we adapt the training setup of Exp.~1 to use only $50$\% of the training samples, namely, all samples with machine parameter settings corresponding to the origin and its $3$ directly connected corners (where only one machine parameter varies). This results in $(4 \times 10)^2=1600$ training inputs when pairing each sample with any other. 
For evaluation, we take all remaining $23 \times 10=230$ samples not used for testing and pair them with the signals resulting from the parameters at the origin as the reference signal, resulting in $(23 \times 10) \times (1 \times 10)=2300$. Again, we repeat the evaluation $6$ times and report the mean figures of merit.

\emph{Experiment 3: Ablation of the latent predictor $P_z$.}
We assess the effect of the latent predictor $P_z$ on the results by
comparing the performance of two model variants: one without $P_z$ (making the model effectively a direct action regressor) and one with it (identical to the model in the previous experiments). The training and test setup are identical to Exp.~1.
This ablation study clarifies whether joint embedding predictive modeling is critical for learning an action-aware representation, leading to improved control.

\emph{Experiment 4: Transfer learning.}
We evaluate the world model's ability, without (4.1) and with (4.2) $P_z$, to learn economically by leveraging prior knowledge.

First, we pre-train the model on dataset D3, forming  training samples by pairing each signal with any other, yielding $460^2=211.6k$ inputs. 
Second, we fine-tune with the setup of Exp.~1, starting from the pre-trained model, to test adaptation from one process to another.


\subsection{Quantitative evaluation}
We aim at evaluating model performance regarding its ability to issue correct control actions, determined by its ability to predict the actions leading from reference signal to observation. This in turn hinges on the model's ability to accurately disentangle the effects of the individual components (change of one machine parameters) of each action. We quantify this by means of three metrics, detailed below: The action direction $\theta$, action magnitude $d$, and overall prediction quality $q$.



\emph{Action direction $\theta$.} 
Does a predicted action point into the correct direction?
We consider the ground truth and predicted action vectors $\boldsymbol{a}$ and $\boldsymbol{\hat{a}}$, 
and calculate the angle $\theta(\boldsymbol{a}, \boldsymbol{\hat{a}}) = \arccos\left( \frac{\boldsymbol{a} \cdot \boldsymbol{\hat{a}}}{\|\boldsymbol{a}\| \cdot \|\boldsymbol{\hat{a}}\|} \right)$ between $\boldsymbol{a}$ and $\boldsymbol{\hat{a}}$ (on a scale between $0$° to $180$°). 
A smaller angle indicates better disentanglement, implying that the model can effectively represent the action's direction in the latent space.  

\emph{Action magnitude $d$.} 
Does a prediction accurately quantify the magnitude of the needed adjustment?
This is answered by the distance between $\boldsymbol{a}$ and $\boldsymbol{\hat{a}}$. 
We compute the absolute distance, where smaller distances indicate more accurate predictions. 
Specifically, 
$d(\boldsymbol{a}, \boldsymbol{\hat{a}})= \sqrt{\sum_{i=1}^n (a_i - \hat{a}_i)^2}$, where $n$ denotes the dimension of the vectors.

\emph{Overall prediction quality $q$.} 
We combine the angle (normalized to [0, 1]) and distance metrics into a single scalar evaluation score to provide a balanced measure of right direction (disentanglement) and magnitude. Specifically, we compute their harmonic mean as $q(\boldsymbol{a}, \boldsymbol{\hat{a}})=\frac{1}{2}\left( \frac{1}{\theta(\boldsymbol{a}, \boldsymbol{\hat{a}})} + \frac{1}{d(\boldsymbol{a}, \boldsymbol{\hat{a}})} \right)^{-1}$. It is well suited as it necessitates the simultaneous minimization of all criteria for good scores.

\emph{2D and 3D analysis.}
Our evaluation employs complementary 2D and 3D analyses to examine the model's action prediction capabilities at different levels of complexity. 
We are first interested to see pairwise comparisons between any $2$ action dimensions individually (2D) for fine-grained analysis  
as this shows how the latent representation responds to changes in individual machine parameters. Specifically, for action dimensions $i$ and $j$, let
$\boldsymbol{a_{i,j}} =[a_i, a_j]$, $\boldsymbol{\hat{a}_{i,j}} = [\hat{a}_i, \hat{a}_j]$, and the 2D angle between them be 
$\theta(\boldsymbol{a_{i,j}}, \boldsymbol{\hat{a}_{i,j}}) = \arccos\left( \frac{\boldsymbol{a(i,j)} \cdot \boldsymbol{\hat{a}(i,j)}}{\|\boldsymbol{a(i,j)}\| \cdot \|\boldsymbol{\hat{a}(i,j)}\|} \right)$.
Then, we calculate the average angle over all possible 2D pairs as 
$\theta_{2D}(\boldsymbol{a}, \boldsymbol{\hat{a}}) = \frac{1}{m} \sum_{i,j \in n, i \neq j} \theta(\boldsymbol{a_{i,j}}, \boldsymbol{\hat{a}_{i,j}})$, where $m$ is the total number of unique pairs (i.e.,  $m=\frac{n(n-1)}{2}=3$ in our case).
The resulting average angle $\theta_{2D}$ serves as a measure of the model's capacity  to disentangle the action dimensions. Correspondingly, 2D distances can be calculated as
$d(\boldsymbol{a_{i,j}}, \boldsymbol{\hat{a}_{i,j}}) = \sqrt{\frac{1}{2}[(a_i - \hat{a}_i)^2 + (a_j - \hat{a}_j)^2]}$
and the average distance over all possible 2D pairs as $d_{2D}(\boldsymbol{a}, \boldsymbol{\hat{a}}) = \frac{1}{m} \sum_{i,j \in n, i \neq j} d(\boldsymbol{a_{i,j}}, \boldsymbol{\hat{a}_{i,j}})$.

All $3$ 2D metrics assess the prediction quality of the individual parameter combinations and the model’s performance in isolating individual parameter changes.
Then, the evaluations are extended to all action dimensions simultaneously (3D), analyzing the model's ability to generalize disentanglement across the full action space. Specifically, we compute $\theta(\boldsymbol{a}, \boldsymbol{\hat{a}})$, $d(\boldsymbol{a}, \boldsymbol{\hat{a}})$ and $q(\boldsymbol{a}, \boldsymbol{\hat{a}})$ by extending the vectors in 3D using the formulas from above. This provides insight on how the model handles multi-parameter changes. 


\section{Results}
\subsection{Exploring the latent space}

\begin{figure}[!t]
    \centering
    \includegraphics[width=0.45\textwidth]{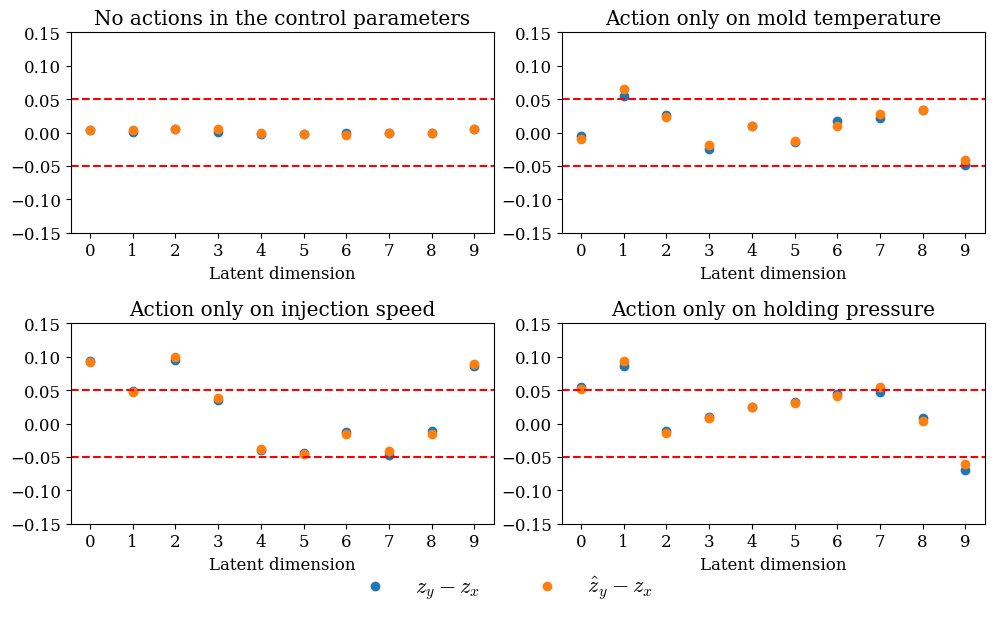}
    \caption{
    Visualization of \emph{differences} between embeddings (either directly as encoded, blue; or as encoded by $Enc$ for $\boldsymbol{x}$ and predicted by $P_z$ for $\boldsymbol{y}$, orange) in the $10$-dimensional embedding space. The red dashed lines indicate a threshold for significantly different activity between the two embeddings. Four different signal pairs are shown: One with a no action (top left), and one each for a change only in mold temperature, injection speed or holding pressure (top right, bottom left and bottom right).
    }
    \label{fig:latent}
\end{figure}

\emph{Action-awareness in the latent representation.}
The model resulting from Exp.~1 on dataset D1 is analyzed regarding properties of its learned latent space. Fig.~\ref{fig:latent} first shows (all $4$ panels) that the differences between the actual latents ($\boldsymbol{z_y} - \boldsymbol{z_x}$) are well aligned with the same differences between the reference's latent and the prediction of the observation's latent ($\boldsymbol{\hat{z}_y} - \boldsymbol{z_x}$). This demonstrates the model's ability to accurately predict activations in latent space based on actions, enabling simulation of the latent space for arbitrary actions.
Moreover, the variation in specific latent dimensions with respect to different actions (all but top-left panel) highlights the interpretability of the latent space and its sensitivity to action-specific behaviors: It appears action-aware. 
For example, as depicted in the bottom left corner of Fig.~\ref{fig:latent}, when action is only taken regarding injection speed, the $0^{th}$, $2^{nd}$ and $9^{th}$ dimension change. On the other hand, when the mold temperature is adjusted, only the $1^{st}$ dimension is significantly affected. These effects are mutually exclusive.

\begin{figure}[!ht]
    \centering
    \includegraphics[width=0.3\textwidth]{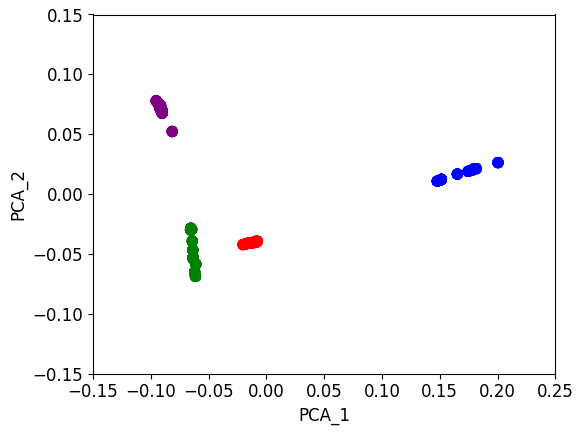}
    \caption{Visualization of embeddings from D1's test set for Exp.~1 in $2$ dimensions using PCA, colored by the actions that led to the underlying signals. Red dots represent embeddings with no action taken, serving as the reference for the other signals; green, blue and purple dots are created by changing only injection speed, molding temperature or holding pressure.  } 
    \label{fig:pca}
\end{figure}

\emph{Clustering in the latent space.}
We map our $10$-dimensional latent space into $2$ dimensions using PCA to gain perceptual insight. Fig.~\ref{fig:pca} shows how representations from Exp.~1 on dataset D1 cluster. It clearly indicates the separation of signals originating from $4$ different machine parameter settings. Such representations can be further used for downstream tasks like anomaly detection and domain shift identification.

\subsection{Action prediction}

The evaluation results of Exp.~1--4 on the experiment-specific test sets from datasets D1 and D2 are presented in Tab.~\ref{tab:experiment_combined}. The experimental setup is briefly summarized in col.~$2$ (with or without $P_z$ or transfer learning, number of machine parameter settings in training set). The next $3$ columns give the final metrics for the 2D evaluations, followed by the last $3$ columns for the 3D evaluations.

\begin{table*}[htpb]
\centering
\tiny
\caption{Evaluation of different experimental setups on experiment-specific test sets of D1 and D2}
\begin{tabular}{l c c c c c c c c c c c c c}

\toprule
\multirow{2}{*}{\textbf{Experiment}} & \multirow{2}{*}{\textbf{Experimental setup}}& \multicolumn{2}{c}{\textbf{Avg. action dir. $\boldsymbol{\theta_{2D}}$}} & \multicolumn{2}{c}{\textbf{Avg. action mag. $\boldsymbol{d_{2D}}$}} & \multicolumn{2}{c}{\textbf{Avg. overall $\boldsymbol{q_{2D}}$}} & \multicolumn{2}{c}{\textbf{Avg. action dir. $\boldsymbol{\theta}$}} & \multicolumn{2}{c}{\textbf{Avg. action mag. $\boldsymbol{d}$}} & \multicolumn{2}{c}{\textbf{Avg. overall $\boldsymbol{q}$}} \\
\cmidrule(lr){3-4} \cmidrule(lr){5-6} \cmidrule(lr){7-8} \cmidrule(lr){9-10} \cmidrule(lr){11-12} \cmidrule(lr){13-14}
 & & \textbf{D1} & \textbf{D2} & \textbf{D1} & \textbf{D2} & \textbf{D1} & \textbf{D2} & \textbf{D1} & \textbf{D2} & \textbf{D1} & \textbf{D2} & \textbf{D1} & \textbf{D2} \\
\midrule
1 (Base) & + latent predictor, - pre-training, 8 settings & 16.13 & \textbf{9.59} & 0.25 & \textbf{0.17} & 0.13 & \textbf{0.08} & 18.22 & \textbf{12.14} & 0.31 & \textbf{0.20} & 0.15 & \textbf{0.10} \\
2 & + latent predictor, - pre-training, 4 settings & 34.00 & 34.00 & 0.50 & 0.50 & 0.14 & 0.14 & 35.00 & 35.00 & 0.61 & 0.61 & 0.15 & 0.15 \\
3 & - latent predictor, - pre-training, 8 settings & 16.56 & 9.94 & 0.25 & \textbf{0.17} & 0.13 & 0.08 & 18.34 & 12.63 & 0.31 & 0.22 & 0.15 & 0.10 \\
4.1 & - latent predictor, + pre-training, 8 settings & 8.68 & 17.85 & 0.14 & 0.25 & 0.07 & 0.14 & 11.13 & 19.64 & 0.17 & 0.31 & 0.09 & 0.16 \\
4.2 & + latent predictor, + pre-training, 8 settings & \textbf{8.53} & 20.56 & \textbf{0.14} & 0.27 & \textbf{0.07} & 0.16 & \textbf{10.61} & 23.89 & \textbf{0.17} & 0.34 & \textbf{0.09} & 0.19 \\

\bottomrule
\end{tabular}
\label{tab:experiment_combined}
\end{table*}

\emph{Evaluating Exp.~1--4 on D1.} 
The following results highlight the impact of pre-training, the latent predictor, and training setting numbers on performance. Best results are observed in Exp.~4.2 (leveraging maximum training data, transfer learning and $P_z$), achieving the lowest 2D and 3D angle errors ($8.53$° and $10.61$°) and the best harmonic mean values. Pre-training (Exp.~4.2 vs. Exp.~1) significantly reduces errors by nearly a factor of two.  The latent predictor further improves performance, for example, when combined with pre-training (Exp.~4.2 vs. Exp.~4.1). 
However, the latent predictor alone (Exp.~1 vs. Exp.~3) shows only a slight improvement ($2-5\%$). Increasing the training settings (Exp.~1 vs. Exp.~2) boosts accuracy, reinforcing that more data leads to better results.


\begin{figure}[!ht]
    \centering
    \includegraphics[width=0.5\textwidth]{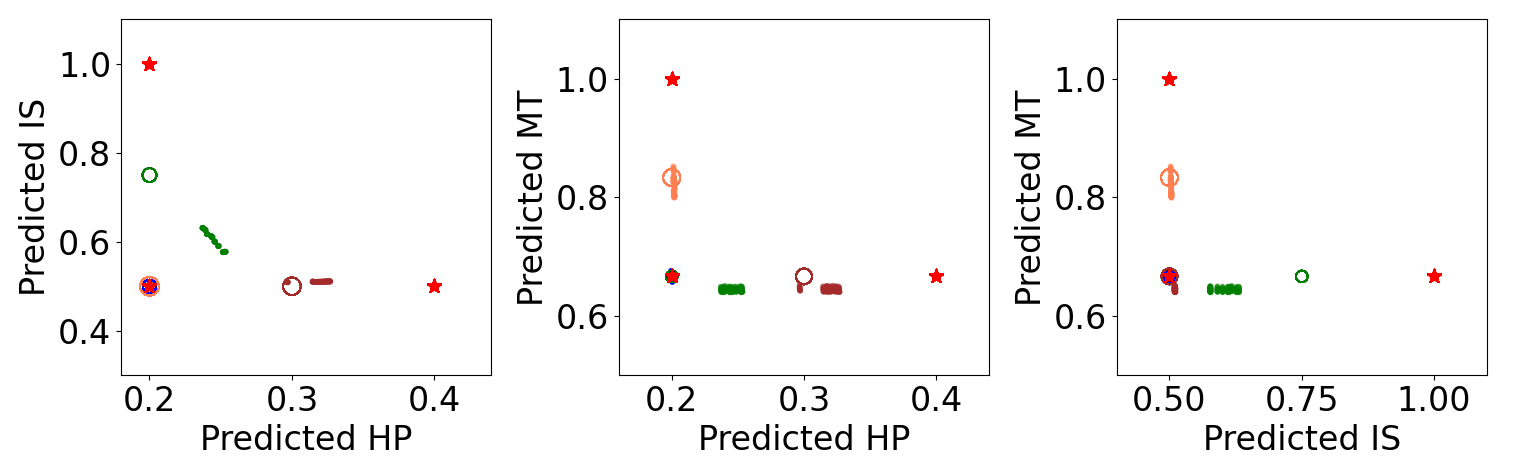}
    \caption{2D pairwise parameter predictions. From left to right: Holding pressure (HP) vs injection speed (IS), HP vs mold temperature (MT), IS vs MT. Red stars represents $3$ training settings, with the bottom left ones serving as reference parameters. Test data with actions changing IS, HP, and MT are labeled in green, purple, and orange, respectively. Hollow cycles represent ground-truth while the predicted actions are indicated by dots.}
    \label{fig:mp_pred}
\end{figure}

\emph{Evaluating Exp.~1--4 on D2.} 
Similar to D1, the model demonstrates poor performance using only $4$ training settings (Exp.~2 vs Exp.~1), as reflected in big angle to and long distance from the ground truth. Increasing the number of settings to $8$ improves performance significantly, showing the importance of geometric context. A slight improvement can be observed when the latent predictor is added (Exp.~3 vs Exp.~1). 
However, pre-training is harmful no matter if latent predictors are absent or not. For instance, Exp.~4.1 shows a degradation in average $\theta$ from $12.14$° (Exp.~1) to $23.89$° and in average $d$ from $0.20$ to $0.34$. This result is understandable and aligns with the discussion in Sec.~\ref{sec:data}: The limited and noisy pre-training data is from a single source, leading to the learned representation not being generalizable to all cases, specifically when the new case is significantly different from the source domain. Then, it causes negative transfer \cite{yan_comprehensive_2024}. 
The best performance is observed in Exp.~1 with latent predictors and no pre-training, achieving the lowest errors across both 2D ($\theta_{2D}$: $9.59$°, $q_{2D}$: $0.08$) and 3D ($\theta$: $12.14$°, $q$: $0.1$). 
Overall, combining latent predictors with sufficient training settings consistently improves performance. However, pre-training is not recommended if the pre-training data significantly differs from the data of the target task.

\emph{Qualitative evaluation.}
The model is designed to accurately predict the deviation of machine parameters relative to the reference, regardless of the scale. 2D pairwise qualitative results on D1 from Exp.~1 are depicted in Fig.~\ref{fig:mp_pred}, comparing predicted actions with ground truth. The figures indicate that the effect of actions regarding holding pressure and mold temperature is well isolated, as the predicted actions are close to the ground truth when these actions are taken. 
However, the effect of injection speed remains entangled with other parameters, indicating that its influence is not as distinctly separable. A key reason is that injection speed and holding pressure are highly correlated in reality, for example, changes in injection speed might influence the material flow which in turn affects pressure.

\section{Conclusions}
We proposed an actionable world model for industrial process control. It shows potential in disentangled latent representation learning and precise action prediction, improving over the state of the art over prior work \cite{yanSDS2024}. This is achieved by creating a novel methodology that generalizes JEPA to learn action-aware representations for arbitrary, factored actions. The results across multiple datasets highlight the importance of the latent predictor, ensuring better disentanglement and improved prediction quality of actions that can be used as process control signals. Additionally, increasing training data (linearly related to the number of settings) enhances alignment with the ground truth and optimizes model's prediction quality. 
However, the results also emphasize necessary caution with respect to pre-training with limited or unrelated data. In some cases, it may lead to performance degradation due to negative transfer from data that do not generalize to the case at hand.

\emph{Limitations.}
The proposed method depends on the availability of machine parameters for task-specific representation learning, acting as external guiding signals for the learning. Accordingly, the method's success relies on the accuracy and reliability of the machine parameters and whether these parameters are truly indicative of the underlying process the model aims to capture. 
Unlike traditional SSL methods (e.g., contrastive learning or I-JEPA), which aim to learn general-purpose representations, this model is tailored to a specific domain or machine setup by incorporating a predictive module. The incorporation of specific actions (control parameters) may limit the model’s ability to generalize to tasks or domains where those parameters are unavailable or differ significantly.

\emph{Future work.}
The model has been applied successfully to injection molding cases with $3$ machine parameters. Next, we want to expand this approach to include process datasets covering more machine parameters.
Furthermore, it would be also interesting to expand this framework for other industrial applications.
Additionally, the effectiveness of transfer learning must be investigated using diverse datasets from different sources, which can be used for downstream fine-tuning. 
A key challenge is the non-bijective relationship between process signals and machine parameters, as the same process signal may correspond to different parameters. 
On a larger scale, actionable world models that learn representations of their percepts with respect to issuing goal-driven behavior are a promising avenue for progress in many other fields of AI, from robotics to autonomous agents.

\bibliographystyle{IEEEtran}
\bibliography{25sds}

\end{document}